# Novel Rough Set based Aggregate Rank-Measure & its Application to Supervised Multi Document Summarization

Nidhika Yadav[a,*], Niladri Chatterjee
[a] *Department of Mathematics, Indian Institute of Technology, Delhi, India.*

**Abstract.** Most problems in Machine Learning cater to classification and the objects of universe are classified to a relevant class. Ranking of classified objects of universe per decision class is a challenging problem. We in this paper propose a novel Rough Set based membership called *Rank-Measure* to solve to this problem. It shall be utilized for ranking the elements to a particular class. It differs from Pawlak's Rough Set based membership function which gives an equivalent characterization of the Rough Set based approximations. It becomes paramount to look beyond the traditional approach of computing memberships while handling inconsistent, erroneous and missing data that is typically present in real world problems. This led us to propose the *aggregate Rank-Measure*. The contribution of the paper is three-fold. Firstly, it proposes a Rough Set based measure to be utilized for numerical characterization of within-class ranking of objects. Secondly, it proposes and establish the properties of *Rank-Measure* and aggregate *Rank-Measure* based membership. Thirdly, we apply the concept of membership and aggregate ranking to the problem of supervised Multi Document Summarization wherein first the important class of sentences are determined using various supervised learning techniques and are post-processed using the proposed ranking measure. The results proved to have significant improvement in accuracy.

## 1. Introduction

Rough Set (Pawlak, 1982) theory is an artificial intelligence technique for decision making under uncertain conditions. Rough Set has wide application areas including but not limited to Financial Prediction (Tay & Shen, 2002), Data Mining (Grzymala-Busse, 2003), Classification, Retrieval (Das-Gupta, 2001, Singh & Dey, 2005), Attribute Reduction (Jensen & Shen, 2007, Wang et al, 2007, Inbarani et al, 2014, Yadav & Chatterjee, 2017), Text Summarization (Yadav & Chatterjee, 2019) to mention few.

The building block of Rough Set based applications is indiscernibility relation. Typically, the objects of universe U are partitioned into equivalence classes based on the knowledge base given in form of attributes, R. The pair (U, R) is called an Information System. The partitions so obtained are also called concepts. When a decision characterization, D of objects of universe is also provided, the triplet (U, R, D) is referred to as decision system. A decision class is a subset of Universe and is represented as a pair of *lower approximation* and *upper approximation*. Lower approximation is the set of concepts that are contained in the decision class while upper approximation is the set of concepts that intersects with the decision class.

Multi-document summarization (Mani and Maybury, 1999) condenses a collection of documents to produce a shortened representative of the documents. With the increase in amount of text data available from various sources Multi-document summarization (MDTS) has become of paramount importance. The MDTS techniques are used in various applications such as summarizing news from multiple sources, using summarizers over the retrieval engines and for providing personalized news feedbacks. Natural languages are complex, unstructured and ambiguous, this research proposes to examine the suitability of the Rough set-based supervised techniques for Multi-document Summarization for its inherent ability to handle inconsistencies present in data.

---

[*]Corresponding author. E-mail: nidhikayadav8@gmail.com

MDTS is characterized as abstractive or extractive based on the criterion of sentences present in the summary. Extractive summary picks up sentences as they are from the source while abstractive summary requires fusion and recreation of original sentences present in documents. Traditional summarization classification techniques require much lower dimensional representation of text as against unsupervised approach. Unsupervised summarization techniques are primarily based on searching for key topics covered in the text without considering the decision class while supervised techniques typically consider the problem as a two-class classification problem given the features of the text. Selecting appropriate features for Multi-document summarization is paramount to lead to efficient classification. The features most popularly involve characteristics of sentences such as sentence position, similarity with title, presence of proper nouns and numeric quantities, cohesion and coherence to mention a few.

In supervised learning one decision class corresponds to the sentence being important from summarization perspective while the other class corresponds to non-relevant sentences. A supervised two class classification algorithm trained on training data can be used to predict the class of the sentence during the testing phase. A large amount of annotated data is made available by Document Understanding Conference (DUC). This data is in the format of abstracts made by human experts. The data has been used in creating labelled classes for training.

The motivation behind the work is firstly to propose a novel Rough Set based membership functions to rank elements of universe for their belongingness to a decision class. Each element of the universe may belong to the decision class $D_i$ with a membership measure called *Rank-Measure*. Here we propose to use the theory of Rough Sets to evaluate the Rank-Measure. Once the ranks are assigned top elements of universe belonging to a particular decision class are selected. A need for such a system is required when a subset of classified decisions is required. One such application area is supervised Multi Document Text Summarization (MDTS). Wherein typically there are two classes viz. relevant vs. non-relevant. The MDTS data is huge and the supervised classes obtained after classification by a Machine Learning algorithm is yields a decision classes of much higher cardinality than is required in the output viz. 10% to 15% of the data.

The major work related to application of Rough Sets to ranking are given as follows. (Greco, Matarazzo & Slowinski, 1999) have proposed multi-criterion decision making using Rough Sets wherein decision rules generated rough approximation of a preference relation is used for ranking. (Zhang et al., 2010) studies object ranking in ordered Information System. Firstly, it transforms the data into binary form. The authors have conducted experiments on university ranking datasets. Here, decision class was used only to compute the importance of an attribute. Further, the decision class represented the numeric rank of the data object. (Naim and Ali, 2011) proposed technique for rank-fusion using Rough Sets. None of these works propose ranking of objects of universe using membership functions as proposed by us.

We also propose to use the popular Rough Sets based classifiers viz. LEM1 (Gyzmala, 2005) and FuzzyRoughNN (Jensen and Cornelis, 2011) for MDTS. The motivation behind it is to unleash the supervised Rough Set based technique for classification while dealing with the vagueness present in natural languages simultaneously for which no previous work has been reported to the best of our knowledge. Further, any classification algorithm for MDTS selects the key sentences according to the classifier and to note that no ranking is performed. The absence of ranking of the sentence in such supervised techniques makes selection of sentences limited to top sentences of corpus which are labelled relevant ignoring the remaining equally competent ones. Hence, we have designed a ranking procedure on the top of a classifier. This ranking is performed on the basis of a Rough Set based membership function. We have evaluated the Rough Set based rankings for the effectiveness of the proposed techniques.

In this research, we have performed extensive analysis and experiments. In this paper we have considered the problem of generic Text Summarization and have tested it on DUC2003 and DUC2005 datasets using DUC2002 datasets as training set. We consider the task of generic summarization and generate 100 words and 250 words extractive summaries for DUC2003 and DUC2005 datasets respectively. We conclude that Rough Set based techniques perform better using the following three criteria. Firstly, its performance is better in terms of various ROUGE scores among most popular supervised techniques which we have tested. Secondly, we learn that the Rough Set based memberships improve the results of almost all the supervised techniques being analyzed in the paper. Further, among the membership based post-processing of all the supervised techniques also Rough Set based algorithms are performing best.

The paper is organized as follows. Section 2 presents the previous work on MDTS. Section 3 gives introduction to Rough Sets and the proposed concepts of aggregate Rank-Measure are defined and explained. Section 4 provide the techniques to apply these concepts to compute extractive summary of text documents. Section 5 discusses the results of experiments conducted on single document Text Summarization viz. DUC2002, DUC2003 and DUC2005 datasets. Finally, Section 6 concludes the paper.

## 2. Previous works

Multi-document Summarization has evolved with time and various techniques were developed for extractive Multi-document Summarization. Some of the key unsupervised and supervised Multi-document Summarization techniques that have been developed so far are discussed below.

Maximum margin relevance (Carbonell and Goldstein, 1998) is a popular technique for query focused Multi-document summarization where relevance to the query is maximized and at the same time redundancy is minimized. Another set of popular algorithms for summarization are centroid based methods (Radev et al., 2000, Hatzivassiloglou, 2001, Radev et al., 2004). A centroid is a set of words that are statistically important to a cluster of documents. Three features which are used to compute the salience of a sentence are centroid value, position of sentence and overlap of sentence with the first sentence. Centroid based methods are efficient for the purpose of summarization and tackle the problem of redundancy by the very formulation of clusters. Also, being an unsupervised technique, there is no requirement for training data. The main disadvantage of the method is that it requires an external parameter (threshold) to form the clusters.

Various supervised techniques such as Neural Networks (Chatterjee, 2018), SVM (Chali et al., 2009), Genetic Algorithms (Chatterjee, 2012), Fuzzy Inference Engine (Kiani et al., 2006, Suanmali et al., 2009), Bayesian Learning (Sharan et al, 2008) have been proposed in literature. Recently Neural Network and GA hybrid technique was proposed (Chatterjee et al., 2018). The features used in this research are as follows: (i) theme similarity, (ii) sentiment factor, (iii) cohesion factor, (iv) readability factor, (v) aggregate similarity and (vi) sentence position. The results proved an improvement over baselines.

G-Flow (Christensen et al., 2013) aims at increasing the coherence in summarized text at the same time choosing sentences with high salience. Coherence represented in terms of coherence relations between segments of text such as elaboration, cause and explanation. The key sentences that cover important concepts are determined with SVM classifier that takes surface level features. A directed graph is used for computing coherence of the extract.

Marujo et al. (2015) proposed two techniques for Multi-document summarization. The first approach is a two-step method where each document is summarized individually followed by the concatenation of all the resulting summaries into a document which is further summarized using the same algorithm. The second method instead of considering all summaries simultaneously, takes them one by one and generating at the kth stage the summary of the first k documents, by concatenating the summary s1,2, .. k-1, the summary of the first k-1 documents with the kth document, and summarizing the concatenated document thus produced. The advantage of these techniques is that it reduces redundancy to a great extent.

## 3. Rough Set

Rough Sets have been extensively used for various applications since its inception in areas of classification, feature selection, data reduction, financial data analysis, information retrieval and many more. Rough Sets been popularly used for supervised decision making given a training dataset. Rough Set based learning from examples often leads to good prediction on the testing data.

Consider a universe $U$ of objects under consideration. Each object of universe is defined by a set of attributes R. Rough Set based techniques are based on the elements which are indistinguishable from each other w.r.t. attributes R. Such elements are said to be in an indiscernibility class. The indiscernibility class of an element x is denoted by [x]. It is defined mathematically as:

$$[x] = \{ y \in U: A[a,x] = A[a,y] \; \forall \; a \in R \}$$

Here $A[a,x]$ is the value of decision table for $a^{th}$ attribute and $x^{th}$ object. The partition induced by an indiscernibility relation on U by R is denoted by U|R. We now define the concept of possible belongingness and certain belongingness of a concept in a set Y based on a set of attributes R. These concepts are referred as the lower approximation and upper approximation respectively and the knowledge is provided in terms of attributes. Mathematically, these are given as follows:

$$\underline{R}Y = \cup \{ Z \in U/R : Z \subseteq Y \}$$
$$\overline{R}Y = \cup \{ Z \in U/R : Z \cap Y \neq \emptyset \}$$

Where Y be a subset of U, $\underline{R}Y$ is the Lower approximation, and $\overline{R}Y$ the Upper Approximations. The boundary region is defined as $BN_R(Y) = \overline{R}Y - \underline{R}Y$. The Rough Set based membership is an equivalent characterization of Rough Set and is defined as follows:

$$\mu_P(x) = \frac{|[X]_P \cap X|}{|[X]_P|}$$

### 3.1. Proposed Rank-Measure

We propose to use Rough Set Rank-Measure using indiscernibility relation for improving the results obtained by supervised text summarization techniques. The problem can be viewed as an element ranking problem wherein a rank is assigned to each object of universe given a decision class. The present work has two broad steps.

The Rough Set based Rank-Measure is used to measure the degree of belongingness of a set X into a concept of a partition. It measures the degree of overlap of the set and the concept of partition to which it belongs. It is defined as follows:

$$\rho_P(x) = \frac{|[x]_P \cap X|}{|X|}$$

Here $[x]_P$ is the equivalence class in which x belongs and X is the decision class corresponding to important sentences. This can be interpreted as the membership of a set of sentences to the decision class of important sentences, given the decision class.

*3.2. Limitations of Rank-Measure function*

There are situations wherein errors in measurements or noise may effect the results. In such cases X may intersect with maximum indiscernibility classes $[x]_{ai}$ without intersecting $[x]_P$. Example 1 below illustrate this case. Where decision class X = {x1, x2} and x1 and x2 differ in one bit only. Still the rank produced by Rank-Measure is 0 while intuitively it should be greater than 0. This led us to define Aggregate Rank-Measure as given below.

*3.3. Proposed Aggregate Rank-Measure*

We now define the aggregate Rank-Measure function on U given the decision class X and a set of attributes P as follows:

$$\rho_P^{Agg}(x) = \frac{\left(\sum_{Pi \in P} \mu_{Pi}(x)\right)}{|P|}$$

The aggregate membership function proposed above caters to the membership of x given the set of important sentences X w.r.t. all the features describing the elements of universe U. As can be seen in Example 1 below the aggregate Rank-Measure of x2 is substantially better than using Rank-Measure only.

|    | a1 | a2 | a3 |
|----|----|----|----|
| x1 | 0  | 1  | 1  |
| x2 | 0  | 1  | 0  |
| x3 | 0  | 0  | 0  |
| x4 | 1  | 1  | 1  |
| x5 | 0  | 1  | 0  |
| x6 | 0  | 2  | 1  |

Table 1

**Example 1.** Consider the Information System given in Table 1. Let X = {x2, x5}

$$\rho_P(x1) = \frac{|\{x1,x6\} \cap \{x2,x5\}|}{|\{x2,x5\}|} = 0$$

$$\rho_P^{Agg}(x1) = \frac{1}{3}\{\frac{|\{x1,x2,x3,x5,x6\} \cap \{x2,x5\}|}{|\{x2,x5\}|}$$

$$+ \frac{|\{x1,x2,x4,x5,x6\} \cap \{x2,x5\}|}{|\{x2,x5\}|} + \frac{|\{x1,x4,x6\} \cap \{x2,x5\}|}{|\{x2,x5\}|}\}$$

$$= \frac{1}{3}\{\frac{2}{2} + \frac{2}{2} + \frac{0}{2}\} = \frac{2}{3} = 0.66$$

It is to be noted that x1 and X differ in one bit only. While $\rho_P(x1)$ is 0, $\rho_P^{Agg}(x1)$ is providing a better estimate of membership of x1 to the set X.

The ranks of all the objects of universe are given in Table 2. Hence the list of objects that belong to decision class X={x2, x5} arranged in ascending order is {x2, x5, x1, x3, x4, x6}.

|    | $\rho_P$ | $\rho_P^{Agg}$ |
|----|----------|----------------|
| x1 | 0        | 0.66           |
| x2 | 1        | 1              |
| x3 | 0        | 0.66           |
| x4 | 0        | 0.33           |
| x5 | 1        | 1              |
| x6 | 0        | 0.33           |

Table 2. Rank-Measure and Aggregate Rank-Measure of Example 1.

The proposed measure can be used for misclassification, missing values and errors in measurements. Here a re-ranking of objects of Universe given the set X is obtained from a supervised or unsupervised learning algorithm. Typically, noise in data occurs in many real-life experiments. The values of Rank-Measures and aggregate Rank-Measures are more symbolic to such datasets.

*3.4. Properties of Rough Set based Rank and Aggregate Rank*

**Proposition 1.** $\rho_P(x) = 1$ if and only if $\rho_P^{Agg}(x) = 1$.
**Proof.** Let $\rho_P^{Agg}(x) = 1$

$$\Rightarrow \frac{\left(\sum_{Pi \in P} \rho_{Pi}(x)\right)}{|P|} = 1, \quad \forall i$$

$\Rightarrow \rho_{Pi}(x) = 1, \quad \forall i$

$\Rightarrow \frac{|[x]_{Pi} \cap X|}{|X|} = 1, \forall i$

$\Rightarrow \forall x \in X, x \in [x]_{Pi}, \forall i$

$\Rightarrow x \in [x]_P$

Hence, $\rho_P(x) = 1$

Conversely, let $\rho_P(x) = 1$

$\Rightarrow \frac{|[x]_P \cap X|}{|X|} = 1$

$\Rightarrow \forall x \in X, x \in [x]_P$

$\Rightarrow \forall x \in X, x \in [x]_{Pi}, \forall i$

$\Rightarrow \rho_{Pi}(x) = 1, \quad \forall i$

Hence, $\rho_P^{Agg}(x) = 1$.

Note though $\rho_P(x) = 0$, may not imply $\rho_P^{Agg}(x) = 0$. The counter example is illustrated in Example 1.

**Proposition 2.** If $\rho_P(x) = 1$ then $\rho_Q(x) = 0$, where $Q \subseteq P$.

Proposition 2 may not hold when where $P \subseteq Q$.

**Proposition 3.** If $\rho_P(x) = 0$ then $\rho_Q(x) = 0$, where $P \subseteq Q$.

**Proposition 4.** If $\rho_P^{Agg}(x) = 1$ then $\rho_Q^{Agg}(x) = 1$ where $Q \subseteq P$.

**Proposition 5.** If $\rho_P^{Agg}(x) = 0$ then $\rho_Q^{Agg}(x) = 0$ where $Q \subseteq P$.

Proposition 4 and 5 may not hold when where $P \subseteq Q$.

**Proposition 6.** If $0 < \rho_P(x) < 1$ then $0 < \rho_P^{Agg}(x) < 1$.
**Proof.** Since $\rho_P(x) < 1$, $\exists\ y \in X$ such that $y \notin [x]_P$. Hence, $\exists\ i \in P$ such that $A[x, i] \neq A[y, i]$ and hence $y \notin [x]_{Pi}$. From this it follows that $\rho_{Pi}^{Agg}(x) \neq 1$.
Also since $\rho_P(x) \neq 0$, $\exists\ y \in X$ such that $y \in [x]_P$. Hence, $\forall\ i \in P$ such that $A[x, i] = A[y, i]$ and hence $y \in [x]_{Pi}$. From this it follows that $\rho_{Pi}^{Agg}(x) \neq 0$. Therefore it follows that $0 < \rho_P^{Agg}(x) < 1$.

**Proposition 7.** If $\rho_P(x) = \rho_P(y) = 1$ then $[x]_P = [y]_P\ for\ x, y \in X$.
**Proof.** $\rho_P(x) = \rho_P(y)$
$\Rightarrow \dfrac{|[x]_P \cap X|}{|X|} = \dfrac{|[y]_P \cap X|}{|X|} = 1$
$\Rightarrow X \subseteq [y]_P$ and $X \subseteq [x]_P$

Also, $[x]_P = \{\ u \in U: A[u, i] = A[x, i]\ \forall\ i\ \}$
$[y]_P = \{\ u \in U: A[u, i] = A[y, i]\ \forall\ i\ \}$
$y \in X$, and $X \subseteq [x]_P$
Hence, $A[x, i] = A[y, i]\ \forall\ i$
And form this it follows that $[x]_P = [y]_P$

In general if $\rho_P(x) = \rho_P(y) \neq 1$, the Proposition may not hold.

The next section discuss the application of aggregate Rank-Measure in the area of supervised Multi-Document Text Summarization.

## 4. Multi-document Text Summarization

*4.1. Phases of MDTS*

The phases for supervised MDTS are given as follows:

(1) Feature Extraction: In this phase statistical features are extracted from the given document cluster. Statistical features are extracted from sentences in the first phase, further a new feature, namely sentiments, as suggested by (Yadav & Chatterjee, 2016) have also been used for summarization.

(2) Tagging Training Data: The training data is assigned the decision class based on ROUGE (Lin, 2001) scores. The automatic labelling is generated using ROUGE of the sentences. It involves assigning the classifying target class of relevant or non-relevant class for each sentence. The annotated textual data in DUC2002 is used for training purpose.

(3) Testing Phase: The features and decision class are given as input to several classifiers. The relevant or non-relevant classes are assigned after the testing phase.

(4) Rough Set Based Membership Computations: The fourth phase involve assigning the proposed Rough Set based membership to each of the relevant sentences which are selected in Step 3.

(5) Summary Creation Phase. This step creates the final summary based on Step 3 and 5.

|          | FuzzyNN | Fuzzy Rough NN | KNN    | Logistic | Naïve Bayes | Neural Network | Random Forest | SVM    | LEM1       |
|----------|---------|----------------|--------|----------|-------------|----------------|---------------|--------|------------|
| ROUGE-1  | 0.2798  | 0.2815         | 0.2794 | 0.2599   | 0.2772      | 0.2726         | 0.2476        | 0.2654 | **0.2952** |
| ROUGE-2  | 0.0482  | 0.0480         | 0.0478 | 0.0447   | 0.0457      | 0.0441         | 0.0412        | 0.0460 | **0.0548** |
| ROUGE-L  | 0.2349  | 0.2369         | 0.2354 | 0.2191   | 0.2338      | 0.2330         | 0.2125        | 0.2240 | **0.2459** |
| ROUGE-SU | 0.0866  | 0.0863         | 0.0853 | 0.0798   | 0.0852      | 0.0829         | 0.0761        | 0.0815 | **0.0926** |

Table 3. Results of DUC2003 for various supervised learning algorithms without any post processing

**ALGORITHM.** Proposed Rough Set based Rank-Measure computation for Post-Processing Supervised Summarization

**Input.** Sentences generated from supervised learning algorithm.
**Output:** Relevancy of Sentence Importance
1. Compute target class D from the input viz. sentences classified as important.
2. Parse the sentences in training data using Stanford parser and compute the word embeddings of the sentences using GLOVE representation of the sentences.
3. Add the sentences and all features to Information System (U, A, D), where U is the universe of sentences, A is the GLOVE features and D is the supervised target class generated from the supervised learning algorithm.
4. For the decision class of important sentences. Find aggregate Rank-Measure of each sentence using the proposed membership given above.
5. Sort sentences selected in descending order of aggregate Rank-Measure of the decision class of important senteces generated by the supervised algorithm according.
6. Generate N word summary from the sorted list.

**End**

Fig. 1. Algorithm to compute Rank-Measure and Extract

## 4.2. Feature Extraction

We describe the features that are considered in this research for Multi-document Summarization based on deep analysis of previously chosen features in various supervised summarization techniques. The features we have chosen for our experiments are the best state of art features

(1) Sentence Position: The position of a sentence in a text document plays a key role in determining its importance level (Kupiec et al., 1995, Yeh et al., 2005, Sharan et al., 2008, Chali et al., 2009). The sentences placed at the start of the text are generally more important than the ones in the interior of other paragraphs. Similar explanation goes

|  | FuzzyNN | Fuzzy Rough NN | KNN | Naïve Bayes | Neural Network | Random Forest | SVM | LEM1 |
|---|---|---|---|---|---|---|---|---|
| **ROUGE-1** | 0.2859 | 0.2863 | 0.2832 | **0.2884** | 0.2689 | 0.2623 | 0.2770 | 0.2796 |
| **ROUGE-2** | 0.0375 | 0.0363 | 0.0353 | 0.0368 | 0.0333 | 0.0337 | 0.0348 | **0.0375** |
| **ROUGE-L** | **0.2386** | 0.2377 | 0.2344 | 0.2382 | 0.2234 | 0.2189 | 0.2304 | 0.2336 |
| **ROUGE-SU** | 0.0870 | 0.0865 | 0.0854 | **0.0871** | 0.0812 | 0.0796 | 0.0668 | 0.0861 |

Table 4. Results of DUC2005 for various supervised learning algorithms without aggregate Rank-Measure

|  | FuzzyNN | Fuzzy RoughNN | KNN | Logistic | Naïve Bayes | Neural Network | Random Forest | SVM | LEMS |
|---|---|---|---|---|---|---|---|---|---|
| **ROUGE-1** | 0.2940 | **0.2956** | 0.2954 | 0.2750 | 0.2858 | 0.2903 | 0.2647 | 0.2789 | 0.2952 |
| **ROUGE-2** | 0.0508 | 0.0517 | 0.0518 | 0.0483 | 0.0476 | 0.0493 | 0.0460 | 0.0490 | **0.0548** |
| **ROUGE-L** | 0.2476 | **0.2492** | 0.2507 | 0.2312 | 0.2392 | 0.2448 | 0.2246 | 0.2352 | 0.2459 |
| **ROUGE-SU** | 0.0920 | 0.0919 | 0.0921 | 0.0856 | 0.0877 | 0.0902 | 0.0829 | 0.0866 | **0.0926** |

Table 5. Results of DUC2003 for various supervised learning algorithms with aggregate Rank-Measure

and are given as follows.

| | FuzzyNN | FuzzyRoughNN | KNN | Logistic | Naïve Bayes | Neural | Random-Forest | SVM | LEMS |
|---|---|---|---|---|---|---|---|---|---|
| **ROUGE-1** | 0.3064 | **0.3095** | 0.3053 | 0.2892 | 0.3058 | 0.2692 | 0.2811 | 0.2899 | 0.3041 |
| **ROUGE-2** | 0.0418 | 0.0404 | 0.0418 | 0.0392 | **0.0428** | 0.0334 | 0.0391 | 0.0377 | 0.0406 |
| **ROUGE-L** | **0.2512** | 0.2505 | 0.2492 | 0.2367 | 0.2494 | 0.2238 | 0.2318 | 0.2366 | 0.2469 |
| **ROUGE-SU** | 0.0941 | 0.0938 | 0.0932 | 0.0883 | **0.0944** | 0.0814 | 0.0873 | 0.0880 | 0.0935 |

Table 6. Results of DUC2005 for various supervised learning algorithms with aggregate Rank-Measure

for sentences at the end of the text. We shall follow the approach given in (Chatterjee et al., 2018), which measures of importance of the i$^{th}$ sentence in a document with N sentences as follows:

$$SP_i = \frac{2(N-i)}{N(N+1)}$$

(2) Sentiment score: Sentiment scores have recently been proposed for text summarization (Chatterjee et al., 2016). The sentiment associated with a text fragment is the opinion that persists in it. Higher sentiment in the text fragment emphasize presence of certain key facts. The neutral sentiment conveys absence of any key subjectivity by the author. In this work we have used SentiWordNet (Baccianella et al., 2010) for computing sentiment score of words present in the sentence. A sentence containing a greater number of positive words is more important that one without any positive words. Hence, in this work a count of number of positive and negative words is taken into consideration. Kiani and Akbarzadeh (2006) and Chali et al. (2009) have use the concept of cue words which is a quite related topic.

(3) Cohesion (Yeh et al., 2005): Cohesion is the property of mutual relatedness of sentences among each other. Cohesion is computed (Chatterjee et al., 2018) using sentence to sentence similarity for each sentence of the collection.

(4) TF-ISF: (Suanmali et al., 2009) have used word frequency as a feature. The sum of tf-isf scores is computed to evaluate the importance of a particular sentence.

(5) Presence of Nouns: Presence of Nouns play a key role in determining the salience of a sentence (Kupiec et al., 1995).

(6) Presence of Numeric Quantities: Presence of numeric quantities also emphasize importance of a sentence (Suanmali et al., 2009).

### 4.3. Rank Computation

We utilize word embeddings, viz. GLOVE (Pennington et al, 2014) to form the decision system used for membership computations. The algorithm is given in Fig. 1.

### 4.4. Supervised Algorithms

We have performed and experimented on supervised learning techniques namely, KNN, FRNN, SVM, Neural Networks, Logistic Regression, LEMS, Naïve Bayes, and Random Forest. The ROUGE scores are measured for without Rank-Measure, with classical membership and with aggre-gate Rank-Measure. We use the implementation of the following which are part of Weka (Hall et al, 2009 and Jensen, 2014) and RSESLIB (Bazan, J. G., & Szczuka, 2000) for Rough Set based algorithms.

| | Fuzzy NN | Fuzzy RoughNN | KNN | Logistic | Naïve Bayes | Neural Network | Random Forest | SVM |
|---|---|---|---|---|---|---|---|---|
| **ROUGE-1** | 5.07 | 4.99 | 5.70 | 5.80 | 3.08 | 6.51 | 6.91 | 5.09 |
| **ROUGE-2** | 5.37 | 7.64 | 8.24 | 8.20 | 4.16 | 11.88 | 11.73 | 6.57 |
| **ROUGE-L** | 5.41 | 5.21 | 6.48 | 5.50 | 2.31 | 5.10 | 5.70 | 5.00 |
| **ROUGE-SU** | 6.22 | 6.50 | 7.91 | 7.30 | 2.89 | 8.83 | 8.87 | 6.27 |

Table 7. Percent increase in DUC2003 results for various supervised learning algorithms with aggregate Rank-Measure.

|  | FuzzyNN | Fuzzy RoughNN | KNN | Naïve Bayes | Neural Networks | Radom-Forest | SVM |
|---|---|---|---|---|---|---|---|
| **ROUGE-1** | 7.17 | 8.12 | 7.82 | 6.03 | 0.12 | 7.18 | 4.68 |
| **ROUGE-2** | 11.59 | 11.48 | 18.36 | 16.37 | 0.30 | 16.19 | 8.44 |
| **ROUGE-L** | 5.32 | 5.42 | 6.36 | 4.71 | 0.20 | 5.92 | 2.72 |
| **ROUGE-SU** | 8.17 | 8.58 | 9.24 | 8.40 | 0.30 | 9.70 | 31.80 |

Table 8. Percent increase in DUC2005 results for various supervised learning algorithms with aggregate Rank-Measure.

## 5. Result and Analysis

We have worked on various datasets viz. DUC2002, DUC2003 and DUC2005 datasets. These datasets contain clusters of multiple news files and each is provided with sample reference summary. We are computing generic summarization of DUC2003 and DUC2005 datasets. DUC2002 Multi-document dataset was used for the purpose of training. In case of DUC2003 multiple 100 word summary are provided while in case of DUC2005 multiple 250 word generic summaries are provided. These summaries were generated by human experts. We use automatic evaluations of summaries generated by our systems and other supervised techniques. The results are evaluated using ROUGE (Lin 2004) score. ROUGE (Recall-Oriented Understudy for Gisting Evaluation), is an n-gram word co-occurrence based technique for automatic evaluation of summaries. Various ROUGE based technique have been developed to compute how effective is a system generated summary and a set of model summaries provided by experts.

The results are presented in Table 3 through 8. In Table 3 results of MDTS for DUC2003 without Rank-Measure are provided. The results are best for Rough Set based LEMS algorithm among all the supervised techniques. Table 4 provides results for DUC2005 datasets without proposed Rank-Measure. Table 5 and 6 provide results for all the supervised techniques with the proposed aggregate Rank-Measure. Table 7 and 8 computes the percentage increase in DUC2003 and DUC2005 with aggregate Rank-Measure used on all the supervised techniques. From Table 7 it is evident that all the supervised techniques for DUC2003 datasets improved on using aggregate Rank-Measure and the increase goes up to 11 percent. It is evident from Table 8 that aggregate Rank-Measure is improving the on all supervised techniques for DUC2005 datasets and the increase goes upto 18 percent.

## 6. Conclusions

This paper presents Rough Set based supervised techniques for Multi-document Summarization. Rough Sets based supervised learning have been used in various applications though its use in supervised text summarization was missing from literature. The present work aims to fill this gap and analyze the applicability of Rough Set based supervised techniques for text summarization. Further, we have evaluated Rough Set based LEMS algorithm to find appropriate subset which can form the extract. LEMS algorithm have not been used for MDTS before this work. Further, we propose novel aggregate Rank-Measure computation for determining the importance of each sentence in the extract. Hence, only top membership sentences were selected for summary generation. Experiments confirm the improvements in result by using the Rough Set based Rank-Measure for sentence importance evaluations. We plan to extend the work using various kinds of rule generation algorithms. Further, we plan to extend the application of aggregate Rank-Measure for other supervised techniques.